\documentclass[10pt, a4paper]{article}
\usepackage{lrec2022} 
\usepackage{multibib}
\usepackage{graphicx}
\usepackage{tabularx}
\usepackage{soul}
\usepackage{xspace}
\usepackage{pgfplots}
\usepackage{pgfplotstable}
\usepackage{enumitem}
\setlist[itemize]{noitemsep, topsep=0pt}
\setlist[enumerate]{noitemsep, topsep=0pt}

\usepackage[autostyle]{csquotes}
\usepackage{numprint}

\definecolorset{cmyk}{GU-}{}{%
  Goethe-Blau,1,0.2,0,0.4;%
  Hellgrau,0.04,0.04,0.05,0.02;%
  Sandgrau,0.12,0.09,0.13,0;%
  Dunkelgrau,0.25,0.25,0.3,0.75;%
  Purple,0.08,1,0.3,0.36;%
  Emo-Rot,0.04,1,0.8,0.07;%
  Senfgelb,0.01,0.25,1,0.05;%
  Gruen,0.62,0.4,0.87,0.09;%
  Magenta,0.08,0.86,0.12,0.12;%
  Orange,0,0.7,1,0.04;%
  Sonnengelb,0,0.12,0.95,0;%
  Helles-Gruen,0.4,0.17,0.81,0.07;%
  Lichtblau,0.8,0,0.06,0.04}
  
\definecolor{HUDarkBlue}{RGB}{16,82,132}
\definecolor{HULightBlue}{RGB}{55,140,195} 
\definecolor{HUBlueText}{RGB}{0,115,186}
\definecolor{HUDarkGrey}{RGB}{206,207,209}
\definecolor{HUMedGrey}{RGB}{219,220,221}
\definecolor{HULightGrey}{RGB}{241,241,241} 
\definecolor{HUDarkOrange}{RGB}{146,76,0} 
\definecolor{LNshadowblue}{RGB}{112,159,203}
\definecolor{LNlightblue}{RGB}{153,196,236}
\definecolor{LNultralightblue}{RGB}{229,240,250}
\definecolor{LNgreen}{RGB}{153,204,51}
\definecolor{GoetheOrange}{RGB}{243,192,57} 
\definecolor{GoetheOrangePlakativ}{RGB}{237,167,45} 
\definecolor{Papier}{RGB}{233,234,192}
\definecolor{agttlightSeminarGrau}{RGB}{219,220,222}
\definecolor{bottomcolor}{rgb}{0.32,0.3,0.38}
\definecolor{middlecolor}{rgb}{0.08,0.08,0.16}
\definecolor{loeweblau}{rgb}{0.33,0.57,0.84}
\definecolor{loewehellblau}{RGB}{219,238,244}
\definecolor{loewegruen}{RGB}{90,162,80}
\definecolor{loeweorange}{RGB}{220,150,55}
\definecolor{DigihumRot}{RGB}{212,0,0}
\definecolor{DigihumBlau}{RGB}{0,94,170}
\definecolor{DigihumHellBlau}{RGB}{221,240,255}
\definecolor{DigihumText}{RGB}{103,102,102}
\definecolor{DigihumHellGrau}{RGB}{214,214,214}
\colorlet{loewegrau}{DigihumText!83!white}
\definecolor{SeminarBlau}{RGB}{0,154,224}
\definecolor{SeminarRot}{rgb}{0.75,0,0}
\definecolor{SeminarGruen}{RGB}{0,128,0}
\definecolor{SeminarOrange}{RGB}{237,167,45} 
\definecolor{SeminarGrau}{rgb}{0.32,0.3,0.38} 
\colorlet{SeminarHellBlau}{SeminarBlau!25!white}
\colorlet{SeminarHellGruen}{SeminarGruen!25!white}
\colorlet{SeminarHellRot}{SeminarRot!25!white}
\colorlet{SeminarHellOrange}{GoetheOrangePlakativ!25!white}
\colorlet{SeminarHellGrau}{SeminarGrau!35!white}
\colorlet{SeminarSehrHellGrau}{SeminarGrau!15!white}
\colorlet{SeminarSehrSehrHellGrau}{SeminarGrau!5!white}

\colorlet{SeminarBlauGruen}{SeminarBlau!50!SeminarGruen}
\colorlet{SeminarGruenBlau}{SeminarBlau!50!SeminarGruen}
\colorlet{SeminarMix}{SeminarBlau!50!SeminarGruen}
\colorlet{SeminarHellMix}{SeminarMix!25!white}
\colorlet{SeminarHellBlauGruen}{SeminarMix!25!white}
\colorlet{SeminarHellGruenBlau}{SeminarMix!25!white}
\colorlet{SeminarRotOrange}{SeminarRot!50!GoetheOrangePlakativ}
\colorlet{SeminarOrangeRot}{SeminarRot!50!GoetheOrangePlakativ}
\colorlet{SeminarHellOrangeRot}{SeminarOrangeRot!25!white}
\colorlet{SeminarHellRotOrange}{SeminarOrangeRot!25!white}
\colorlet{SeminarRotGruen}{SeminarRot!50!SeminarGruen}
\colorlet{SeminarGruenRot}{SeminarRot!50!SeminarGruen}
\colorlet{SeminarRotBlau}{SeminarRot!50!SeminarBlau}
\colorlet{SeminarBlauRot}{SeminarRot!50!SeminarBlau}
\colorlet{SeminarHellRotBlau}{SeminarRotBlau!25!white}
\colorlet{SeminarHellBlauRot}{SeminarRotBlau!25!white}
\colorlet{SeminarHellRotGruen}{SeminarRotGruen!25!white}
\colorlet{SeminarHellGruenRot}{SeminarRotGruen!25!white}
\colorlet{SeminarOrangeGruen}{SeminarOrange!50!SeminarGruen}
\colorlet{SeminarGruenOrange}{SeminarOrange!50!SeminarGruen}
\colorlet{SeminarHellGruenOrange}{SeminarOrangeGruen!25!white}
\colorlet{SeminarHellOrangeGruen}{SeminarOrangeGruen!25!white}
\colorlet{SeminarOrangeBlau}{SeminarOrange!50!SeminarBlau}
\colorlet{SeminarBlauOrange}{SeminarOrange!50!SeminarBlau}
\colorlet{SeminarHellBlauOrange}{SeminarOrangeBlau!25!white}
\colorlet{SeminarHellOrangeBlau}{SeminarOrangeBlau!25!white}
\colorlet{SeminarSehrHellRot}{SeminarRot!15!white}
\colorlet{SeminarSehrHellGrau}{SeminarGrau!15!white}
\colorlet{SeminarSehrHellGruen}{SeminarGruen!15!white}
\colorlet{SeminarSehrHellBlau}{SeminarBlau!15!white}
\colorlet{SeminarSehrHellOrange}{SeminarHellOrange!15!white}
\colorlet{SeminarOrangeGrau}{SeminarOrange!50!SeminarGrau}
\colorlet{SeminarGrauOrange}{SeminarOrange!50!SeminarGrau}
\colorlet{SeminarRotGrau}{SeminarRot!50!SeminarGrau}
\colorlet{SeminarGrauRot}{SeminarRot!50!SeminarGrau}
\colorlet{SeminarBlauGrau}{SeminarBlau!50!SeminarGrau}
\colorlet{SeminarGrauBlau}{SeminarBlau!50!SeminarGrau}
\colorlet{SeminarGruenGrau}{SeminarGruen!50!SeminarGrau}
\colorlet{SeminarGrauGruen}{SeminarGruen!50!SeminarGrau}
\colorlet{SeminarHellGrauRot}{SeminarGrauRot!25!white}
\colorlet{SeminarHellRotGrau}{SeminarGrauRot!25!white}
\colorlet{SeminarHellGrauBlau}{SeminarGrauBlau!25!white}
\colorlet{SeminarHellBlauGrau}{SeminarGrauBlau!25!white}
\colorlet{SeminarHellGrauGruen}{SeminarGrauGruen!25!white}
\colorlet{SeminarHellGruenGrau}{SeminarGrauGruen!25!white}
\colorlet{SeminarHellGrauOrange}{SeminarGrauOrange!25!white}
\colorlet{SeminarHellOrangeGrau}{SeminarGrauOrange!25!white}

\usepackage{listings}

\lstdefinelanguage{XML}
{
  morestring=[b]",
  morestring=[s]{>}{<},
  morecomment=[s]{<?}{?>},
  stringstyle=\color{black},
  identifierstyle=\color{GU-Goethe-Blau},
  keywordstyle=\color{GU-Lichtblau},
  morekeywords={xmlns,version,type}
}

\pgfplotsset{compat=newest}

\usepackage{tikz}
\usepackage{tikzsymbols}
\usetikzlibrary{babel}
\usetikzlibrary{arrows}
\usetikzlibrary{arrows.meta}
\usetikzlibrary{automata}
\usetikzlibrary{positioning}
\usetikzlibrary{matrix}
\usetikzlibrary{fit}
\usetikzlibrary{backgrounds}
\usetikzlibrary{shadows}
\usetikzlibrary{shapes.multipart}
\usetikzlibrary{shapes.misc}
\usetikzlibrary{shapes.geometric}
\usetikzlibrary{shapes.arrows}
\usetikzlibrary{shapes.symbols}
\usetikzlibrary{shapes.callouts}
\usetikzlibrary{er}
\usetikzlibrary{shadings}
\usetikzlibrary{topaths}
\usetikzlibrary{chains}
\usetikzlibrary{decorations.pathreplacing}
\usetikzlibrary{decorations.pathmorphing}
\usetikzlibrary{patterns}
\usetikzlibrary{calc}
\usetikzlibrary{mindmap}
\usetikzlibrary{through}

\usepackage{fontawesome}

\usepackage{titlesec}
\titleformat{\section}{\normalfont\large\bfseries\center}{\thesection.}{1em}{}
\titleformat{\subsection}{\normalfont\SmallTitleFont\bfseries\raggedright}{\thesubsection.}{1em}{}
\titleformat{\subsubsection}{\normalfont\normalsize\bfseries\raggedright}{\thesubsubsection.}{1em}{}
\renewcommand\thesection{\arabic{section}}
\renewcommand\thesubsection{\thesection.\arabic{subsection}}
\renewcommand\thesubsubsection{\thesubsection.\arabic{subsubsection}}

\usepackage{epstopdf}
\usepackage[utf8x]{inputenc}

\usepackage{hyperref}
\usepackage{xstring}

\newcommand{\GerParCor}{\textsc{GerParCor}\xspace}
\newcommand{\TI}{\textsc{TextImager}\xspace}

\usepackage{color}

\title{German Parliamentary Corpus (\GerParCor)}

\name{Giuseppe Abrami, Mevlüt Bagci, Leon Hammerla, Alexander Mehler} 

\address{Goethe University Frankfurt\\
         Robert-Mayer-Straße 10, 60325 Frankfurt am Main \\
         \{abrami, bagci, mehler\}@em.uni-frankfurt.de, hammerla@stud.uni-frankfurt.de
         }

\abstract{
Parliamentary debates represent a large and partly unexploited treasure trove of publicly accessible texts. 
In the German-speaking area, there is a certain deficit of uniformly accessible and annotated corpora covering all German-speaking parliaments at the national and federal level.
To address this gap, we introduce the \textit{German Parliament Corpus} (\GerParCor).
\GerParCor is a genre-specific corpus of (predominantly historical) German-language parliamentary protocols from three centuries and four countries, including state and federal level data.
In addition, \GerParCor contains conversions of scanned protocols and, in particular, of protocols in Fraktur converted via an OCR process based on \textsc{Tesseract}.
All protocols were preprocessed by means of the NLP pipeline of spaCy3 and automatically annotated with metadata regarding their session date. 
\GerParCor is made available in the XMI format of the UIMA project.
In this way, \GerParCor can be used as a large corpus of historical texts in the field of political communication for various tasks in NLP.
\\\newline  }

\begin{document}

\pgfplotstableread[col sep=&, header=true]{
parliament & Sitzungen & Sentence & Token
BadenWuertemberg & 412 & 2494970 & 28365464
Bayern & 2377 & 9191955 & 116914415
Berlin & 582 & 3954101 & 46981044
Brandenburg & 442 & 2460840 & 31439980
Bremen & 1070 & 4338171 & 61396356
Bundesrat & 1008 & 2441772 & 31999748
Bundestag & 3719 & 16286016 & 258521349
Hamburg & 586 & 2256178 & 31294553
Hessen & 1297 & 5692122 & 72994750
Liechtenstein & 504 & 2516530 & 30927117
MeckPom & 659 & 3267241 & 45320645
Niedersachsen & 1109 & 6570416 & 82367685
NordrheinWestfalen & 2041 & 8939350 & 115581074
Oesterreich & 3606 & 16282052 & 228214975
RheinlandPfalz & 1562 & 5584254 & 75178248
Saarland & 876 & 3273321 & 48950664
Sachsen & 690 & 4004190 & 52404321
SachsenAnhalt & 607 & 3578857 & 45083355
SchleswigHolstein & 1776 & 6918739 & 87739660
Schweiz & 1194 & 1548029 & 26203941
Thueringen & 761 & 3404991 & 49281475
Reichstag & 1970 & 3086888 & 60023446
Reichstag\_Empire & 2183 & 4744901 & 82456344
Weimar\_Republic & 1331 & 2887216 & 44389348
ThirdReich & 19 & 14704 & 233421
}\parliamentData

\maketitleabstract

\section{Introduction}\label{sec:introduction} 
The creation of language resources that are fully annotated in an optimal way is a major issue which consumes a lot of time and effort.
Nevertheless, in the current era, with increasing digitization and open access strategies, new treasures of corpora can be unearthed.
This includes parliamentary documents, which are available in various types:

\begin{itemize}
    \item \textbf{Plenary protocols}:
    Plenary protocols are stenographic documentations of the plenary session, including speeches, comments and other contributions such as applause.
    In the plenary protocols there are references to printed matters which are being debated.
    
    \item \textbf{Printed matter}:
    All processes which are dealt with in a parliament are referred to as printed matter. These can be draft bills, proposals, reports or questions.
    \begin{itemize}
        \item \textbf{Minor Questions}:
        Members of a parliament may ask their government Minor Questions, which the government must answer and publish in a timely manner.
        
        \item \textbf{Major Question}:
        In addition, Members of Parliament can use a Major Question to request information and clarification from the government on political issues and facts.
        At least in the German Bundestag, the government's answer can be discussed publicly in the plenary session.    
    \end{itemize}
    
    \item \textbf{Committee protocols}:
    Most parliaments discuss issues beforehand in individual committees, which then (among other things) prepare proposals for the plenum. 
    These meetings are usually open to the public and are also minuted.

\end{itemize}
Currently, the latter documents are not yet fully available, which has several reasons: 
many of them are not accessible via a direct path (API), only as scanned images, or not available at all because they have not been digitized.
Since not all of the above-mentioned types of documents are equally available from all German-speaking parliaments, \GerParCor includes only the plenary protocols on a national and federal level in order to create as broad a German parliamentary corpus as possible.
For the distributed processing this corpus, we used \TI~\cite{Hemati:Uslu:Mehler:2016} which utilized spaCy3\footnote{\url{https://spacy.io/}} for NLP-related preprocessing. 
Using spaCy3, we executed the following preprocessing pipeline to enrich \GerParCor with linguistic annotations:
tokenization, 
sentence recognition,
PoS tagging,
lemmatization,
named entity recognition,
morphology recognition and
dependency parsing.

We make all of the annotated documents available using UIMA~\cite{Ferrucci:et:al:2009} and the XMI format.
In addition, for each document, we extract metadata from the documents and add it to the XMI files based on UIMA -- this includes the session date, location, and title, if available.
In this way, \GerParCor enables a time-related analysis of parliamentary text data.
The final corpus, \GerParCor, is available via GitHub (\url{https://github.com/texttechnologylab/GerParCor}).

\section{Related Work} 
Several German-language parliamentary corpora already exist, although some are not primarily based on plenary sessions. 
\newcite{Barbaresi:2018} collects speeches by the German President, the President of the Bundestag, the German Chancellor and the Foreign Minister from the years 1982--2020.
Another collection of tokenized parliamentary debates of the German Bundestag between 1998 and 2015 is presented by \newcite{Truan:2019}.
The \textit{GermaParl} corpus makes available plenary debates between February 1996 and December 2016~\cite{Blaettle:Blessing:2018}.
For the National Council in Austria, \newcite{Wissik:Pirker:2018} created a parliamentary corpus for the years 1996--2016.
For Austria, there is also a corpus of plenary debates from 2013--2015 \newcite{Sippl:et:al:2016}, which was processed using Stanford Tagger.
\textit{ParlSpeech V2}~\cite{Rauh:Schwalbach:2020} contains the parliamentary protocols of the national chambers of Austria, Germany, Denmark and other countries for several periods between 21 and 32 years.
The \textit{DeuParl} corpus of \cite{Kirschner:et:al:2021} contains the plenary minutes of the Reichstag and the Bundestag, in total from 1867 to June 2021.

Since there is no complete corpus of protocols of the national parliaments for Austria, Switzerland, Liechtenstein, or Germany, which would also be constantly updated to include the ever new protocols, we generated \GerParCor to fill this gap.
To round off this task, \GerParCor also contains the minutes of the German federal parliaments.
In this way, a very large corpus of genre-specific (predominantly historical) German-language texts from three centuries from different countries and different political levels is created (in future work we plan to include the minutes of the GDR People's Chamber).

\section{Corpus Building} 

We downloaded all the parliamentary speeches available online to collect the texts of \GerParCor.
We used the APIs of the individual parliaments for this purpose, although this could not be done in a uniform manner.
In some cases, parliaments do not even have an API, but only a website that offers their minutes as downloads, separated by session, often mixing minutes and other material, as described in Section~\ref{sec:introduction}.
Only a few parliaments, such as the Bundestag, offer complete archives for past periods for download.
In addition, the available plenary minutes can often only be downloaded individually, with interfaces differing between parliaments.
Although there is a joint project of the German state parliaments\footnote{\url{https://www.parlamentsspiegel.de}}, only a few of the protocols are available there.
As a consequence, we developed a \textit{separate} download function for each state parliament.
The software is available via GitHub.

\begin{figure}[h!]
\begin{tikzpicture}
    \begin{axis}[
            ybar,
            xtick=data,
            axis y line*=left, 
            axis x line*=bottom,
            enlarge x limits=0.03,
            xticklabels from table={\parliamentData}{parliament},
            xticklabel style = {rotate=90,anchor=east},
            ymin=0,
            bar width=.2cm,
            nodes near coords,
            every node near coord/.append style={rotate=90, anchor=west, scale=0.9}
        ]
        \addplot table[x expr=\coordindex, y=Sitzungen]{\parliamentData};
    \end{axis}
\end{tikzpicture}
    \caption{Number of sessions.}
    \label{fig:sessions}
\end{figure}
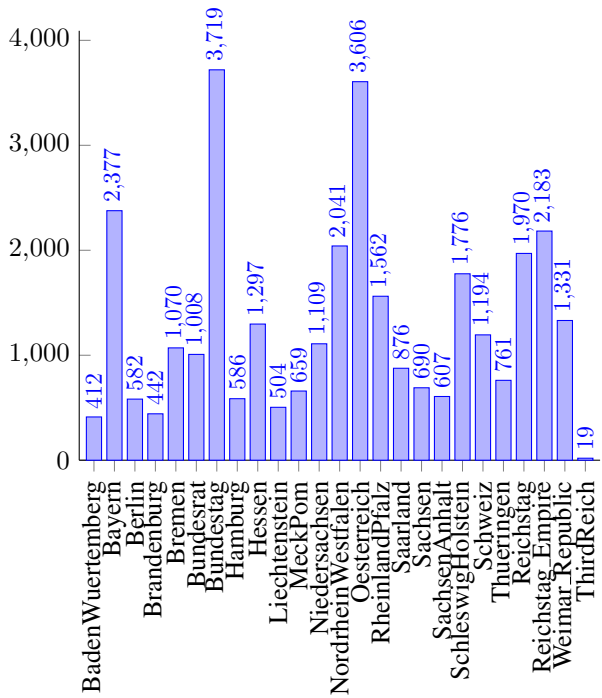

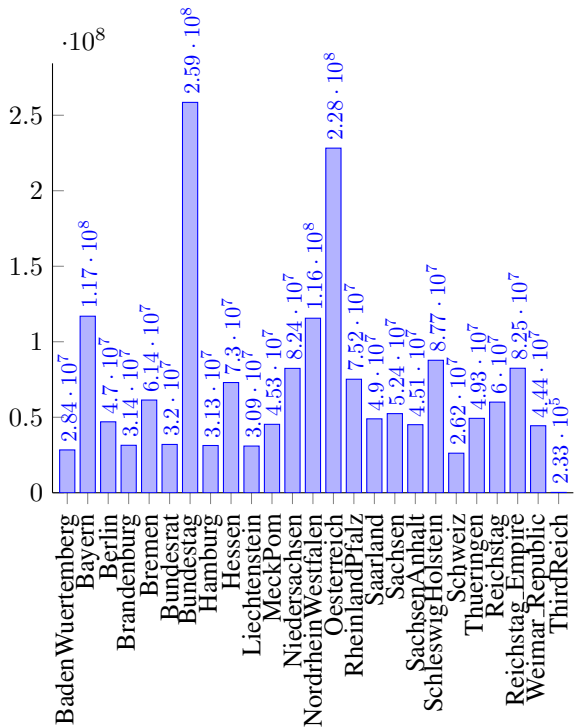
\begin{figure}[h!]
\begin{tikzpicture}
    \begin{axis}[
            ybar,
            xtick=data,
            axis y line*=left, 
            axis x line*=bottom,
            enlarge x limits=0.03,
            xticklabels from table={\parliamentData}{parliament},
            xticklabel style = {rotate=90,anchor=east},
            ymin=0,
            bar width=.2cm,
            nodes near coords,
            every node near coord/.append style={rotate=90, anchor=west, scale=0.85}
        ]
        \addplot table[x expr=\coordindex, y=Token]{\parliamentData};
    \end{axis}
\end{tikzpicture}
    \caption{Number of tokens in the parliaments protocols.}
    \label{fig:tokens}
\end{figure}

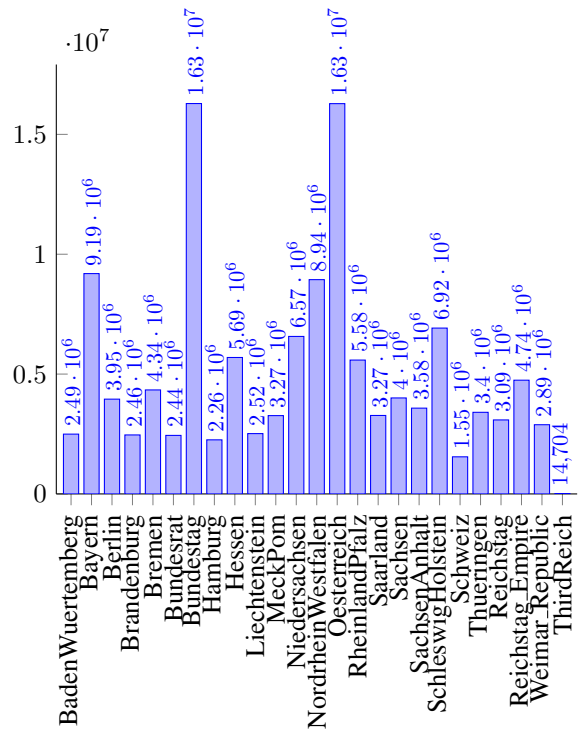
\begin{figure}[h!]

\begin{tikzpicture}
    \begin{axis}[
            ybar,
            xtick=data,
            axis y line*=left, 
            axis x line*=bottom,
            enlarge x limits=0.03,
            xticklabels from table={\parliamentData}{parliament},
            xticklabel style = {rotate=90,anchor=east},
            ymin=0,
            bar width=.2cm,
            nodes near coords,
            every node near coord/.append style={rotate=90, anchor=west, scale=0.85}
        ]
        \addplot table[x expr=\coordindex, y=Sentence]{\parliamentData};
    \end{axis}
\end{tikzpicture}
    \caption{Number of sentences in the parliaments protocols.}
    \label{fig:sentences}
\end{figure}

Some protocols were not available online, but could be made available thanks to the support of the Stenographic Services of the Saarland Parliament, the Bremen Parliament, as well as the Parliament of Rheinland-Pfalz.
However, the plenary minutes of the Niedersachen State Parliament of the 1st to 9th legislative periods were not available in digital form and could not be digitized.
An overview of the automatically recorded protocols of the respective parliaments can be found in Table~\ref{tab:table_dates}. 
The distribution of the corresponding parliamentary sessions is shown in Figure~\ref{fig:sessions} to Figure~\ref{fig:sentences}.

Depending on the dissemination method, the individual protocols were downloaded individually or as a package and preprocessed using spaCy3 \cite{Honnibal:et:al:2020} via \TI~\cite{Hemati:Uslu:Mehler:2016}.
We used \TI because the amount of data required distributed processing, as enabled by \TI.
We additionally extracted metadata from the protocols and annotated this data as instances of the class \textsc{DocumentAnnotation}.
Besides a possible subtitle that contains the legislative period, this \textsc{DocumentAnnotation} also contains the date of the protocol.
A sample XMI annotation is shown in Figure~\ref{fig:xmi}.

\begin{table}[h]
    \begin{tabularx}{\linewidth}{X|l}

\textbf{Parliament}	 & 	 \textbf{Period} \\ \hline
\multicolumn{2}{c}{\textbf{Germany}} \\ \hline 	
Reichstag (North German Union / Zollparlamente)	 &  1867-02-25--1895-05-24 \\ \hline	
Reichstag (German Empire)  &     1895-03-12--1918-10-26 \\ \hline
Weimar Republic  &  1919-02-06--1932-09-12  \\ \hline
Third Reich &  1933-21-03--1942-04-24  \\ \hline
Bundestag  	 & 	 1949-07-09--2021-07-09  \\ \hline
Bundesrat  	 & 	 1949-07-09--2021-08-10  \\ \hline
\multicolumn{2}{c}{\textbf{German Regional Parliaments}} \\ \hline 
Berlin 	 & 	 1989-04-02--2021-09-16  \\ \hline
Bremen 	 & 	 1995-04-07--2021-09-16  \\ \hline
Hamburg 	 & 	 1997-10-08--2021-03-11  \\ \hline
Baden-Württemberg  	 & 	 1984-06-05--2021-09-29  \\ \hline
Bayern 	 & 	 1946-12-16--2021-10-14  \\ \hline
Brandenburg  	 & 	 1990-10-16--2021-08-27  \\ \hline
Hessen  	 & 	 1947-02-04--2021-09-29  \\ \hline
Mecklenburg-Vorpommern  	 & 	 1990-10-26--2021-06-11  \\ \hline
Niedersachsen  	 & 	 1982-06-22--2021-09-15  \\ \hline
Nordrhein-Westfalen  	 & 	 1947-05-21--2021-10-08  \\ \hline
Rheinland-Pfalz  	 & 	 1947-07-24--2021-09-22  \\ \hline
Saarland  	 & 	 1959-07-23--2021-09-15  \\ \hline
Sachsen  	 & 	 1990-10-27--2021-11-18  \\ \hline
Sachsen-Anhalt  	 & 	 1990-10-28--2021-09-17  \\ \hline
Schleswig-Holstein  	 & 	 1946-02-26--2021-02-11  \\ \hline
Thüringen  	 & 	 1990-10-25--2021-11-19  \\ \hline
\multicolumn{2}{c}{\textbf{Liechtenstein}} \\ \hline
Landtag Fürstentums Liechtenstein  	 & 	 1997-03-13--2021-11-06  \\ \hline
\multicolumn{2}{c}{\textbf{Austria}} \\ \hline
Nationalrat (AT) 	 & 	 1918-10-21--2021-05-17  \\ \hline
\multicolumn{2}{c}{\textbf{Switzerland}} \\ \hline
Nationalrat (CH) 	 & 	 1999-06-12--2021-12-09  
\end{tabularx}
    \caption{
    Parliamentary protocols of regional and national parliaments included in \GerParCor.
    }
    \label{tab:table_dates}
\end{table}

\section{OCR} 
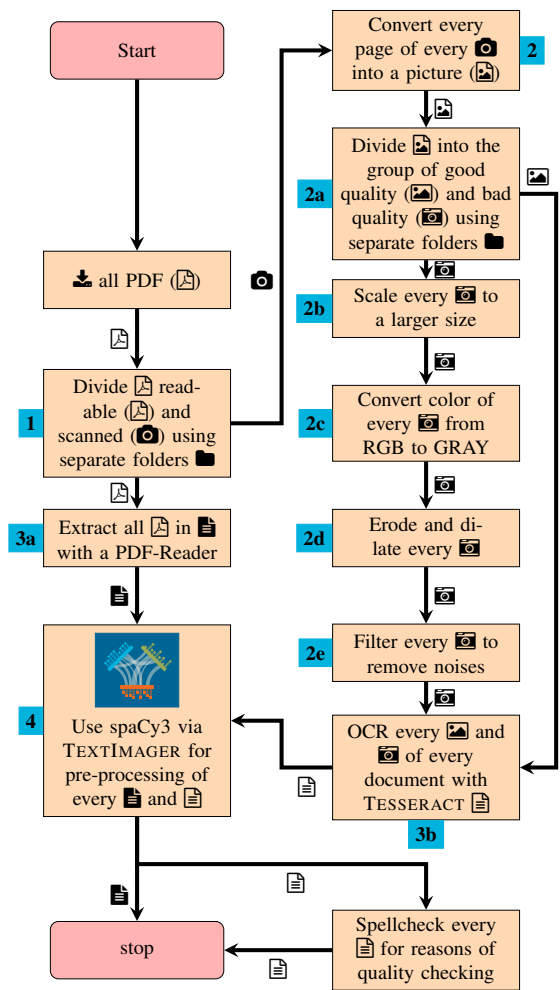
\begin{figure}[h!]

\resizebox{\linewidth}{!}{%

        \tikzstyle{startstop} = [rectangle, rounded corners, minimum width=3cm, minimum height=1cm,text centered, draw=black, fill=red!30]
        \tikzstyle{io} = [trapezium, trapezium left angle=70, trapezium right angle=110, minimum width=3cm, text width=3cm, minimum height=1cm, text centered, draw=black, fill=blue!30]
        \tikzstyle{process} = [rectangle, minimum width=3cm, minimum height=1cm, text centered, text width=3cm, draw=black, fill=orange!30]
        \tikzstyle{decision} = [diamond, minimum width=3cm, minimum height=1cm, text centered, draw=black, fill=green!30]
        \tikzstyle{arrow} = [thick,->,>=stealth, line width=2pt]
        \tikzstyle{nodeLabel} = [rectangle, draw=GU-Lichtblau, fill=GU-Lichtblau,font=\bfseries]

        \begin{tikzpicture}[node distance=2cm]
        \node (start) [startstop] {Start};
        \node (download) [process, below of=start, yshift=-2cm] {\faDownload ~all PDF (\faFilePdfO) };
        \node (divide) [process, below of=download, yshift=-0.5cm,label={[nodeLabel]west:1}] {Divide \faFilePdfO\xspace readable (\faFilePdfO) and scanned (\faCamera) using separate folders \faFolder};
        \node (readable)[process, below of=divide, label={[nodeLabel]west:3a}] {Extract all \faFilePdfO ~in \faFileText\xspace with a PDF-Reader};
        
        \node (convert) [process, right of=start, xshift= 3cm, label={[nodeLabel]east:2}]{Convert every page of  every \faCamera\xspace into a picture (\faFilePictureO)};
        \node (dividefrak) [process, below of=convert, yshift=-0.5cm, label={[nodeLabel]west:2a}]{Divide \faFilePictureO\xspace into the group of good quality (\faPhoto) and bad quality (\faCameraRetro) using separate folders \faFolder};
        \node (resize) [process, below of=dividefrak, label={[nodeLabel]west:2b}] {Scale every \faCameraRetro ~to a larger size};
        \node (gray) [process, below of=resize, label={[nodeLabel]west:2c}] {Convert color of every \faCameraRetro ~from RGB to GRAY};
        \node (ero_dilate) [process, below of=gray, label={[nodeLabel]west:2d}] {Erode and dilate every \faCameraRetro};
        \node (remove_noise) [process, below of=ero_dilate, label={[nodeLabel]west:2e}] {Filter every \faCameraRetro ~to remove noises};
        
        \node (tesseract) [process, below of=remove_noise, label={[nodeLabel]south:3b}]{OCR every \faPhoto ~and \faCameraRetro\xspace of every document with \textsc{Tesseract} \faFileTextO};
        
        \node (spacyTI) [process, below=1cm of readable, label={[nodeLabel]west:4}] {\includegraphics[width=1.5cm, page=30]{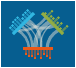}
        
        Use spaCy3 via \mbox{\TI} for pre-processing of every \faFileText\xspace and \faFileTextO};
        \node (stop) [startstop, below of=spacyTI, yshift=-2cm] {stop};
        \node (spell) [process, right of=stop, xshift=3cm, label={[nodeLabel]south:4a}] {Spellcheck every \faFileTextO\ for reasons of quality checking};
        
        \draw [arrow] (start)--(download);
        \draw [arrow] (download)--node[left]{\faFilePdfO}(divide);
        \draw [arrow] (divide)--node[left]{\faFilePdfO}(readable);
        \draw [arrow] (divide)-|node[left, near end]{\faCamera}+(+2.5,5)|-(convert);
        \draw [arrow] (readable)--node[left]{\faFileText}(spacyTI);
        \draw [arrow] (convert)--node[right]{\faFilePhotoO}(dividefrak);
        \draw [arrow] (dividefrak)--node[right]{\faCameraRetro }(resize);
        \draw [arrow] (resize)--node[right]{\faCameraRetro}(gray);
        \draw [arrow] (gray)--node[right]{\faCameraRetro}(ero_dilate);
        \draw [arrow] (ero_dilate)--node[right]{\faCameraRetro}(remove_noise);
        \draw [arrow] (remove_noise)--node[right]{\faCameraRetro}(tesseract);
        \draw [arrow] (dividefrak)-|node[above, near start]{\faPhoto}+(+2.25, -2.5)|-(tesseract);
        \draw [arrow] (tesseract)-|+(-2.5, 0.25)node[below, near start]{\faFileTextO}|-(spacyTI);
        \draw [arrow] (spacyTI)--node[left, near end]{\faFileText}(stop);
        \draw [arrow] (spacyTI)|-+(+0.5, -2.5)-|node[below, near start]{\faFileTextO}(spell);
        \draw [arrow] (spell)--node[below]{\faFileTextO}(stop);

        \end{tikzpicture}
}
    \caption{Workflow of \GerParCor's OCR process including NLP preprocessing.}
    \label{fig:ocr_schema}
\end{figure}

Some parliamentary minutes were only available as scanned copies, so they had to be pre-processed with OCR (see Tab.~\ref{tab:table_spellchecking}).
Moreover, some of these scans are only available in Fraktur.
To convert these scans into text, \textit{Optical Character Recognition} (OCR) was performed using \textsc{Tesseract}~\cite{Kay:2007} from Google.
\textsc{Tesseract} provides various language models for text recognition, including German {Fraktur}.\footnote{\url{https://tesseract-ocr.github.io/tessdoc/Data-Files-in-different-versions.html}}
To perform OCR, the individual PDF documents must be converted into images page by page as shown in the workflow in Figure~\ref{fig:ocr_schema} and described by the following procedure:

\begin{enumerate}
    \item Divide all downloaded PDF documents into readable PDF documents (\faFilePdfO) and into scanned documents (\faCamera).
    \item Convert every page of every scanned document (\faCamera) (python library: pdf2image~\cite{pdf2image:2017}).
    \begin{enumerate}
        \item 
        Divide the documents into good and poor quality scans; if there are only good quality scans, proceed to point 3.
        \item[(b-e)] 
        For each image of the group of bad scans: rescale, convert the color from RGB to gray, erode, dilate and remove/reduce noise with a filter (Python library: OpenCV~\cite{opencv:2017}).
    \end{enumerate}
    \item Text-extraction:
    \begin{enumerate}
        \item Extract the text of every readable PDF document with a PDF extractor (python library: textract\footnote{\url{https://textract.readthedocs.io/en/stable/}}).
        \item Extract the text of every scanned document from converted images using \textsc{Tesseract}.
    \end{enumerate}
    \item NLP-Processing of every text extraction using spaCy 3 via \TI.
    \item Check OCR output quality using a spellchecker (\textit{SymSpell}~\cite{wolfgarbe:2017}).
\end{enumerate}


%
GeeksforGeeks have a good example to extract the text of an PDF document, which we used as a basis for our code \footnote{\texttt{\href{https://www.geeksforgeeks.org/python-reading-contents-of-pdf-using-ocr-optical-character-recognition}{https://www.geeksforgeeks.org/python\-reading-contents-of-pdf-using-ocr-optical\-character-recognition}}}.
To this end, we removed each converted image and used multithreading to speed up the extraction.
By default, \textsc{Tesseract} uses four cores to extract text from images\footnote{\url{https://tesseract-ocr.github.io/tessdoc/FAQ.html}}
Thus, there are two alternatives to prevent overthreading:
\begin{enumerate}
    \item Change the number of cores for text extraction from four to one and start the application with multithreading.
    \item Divide the number of existing threads by four, round the result and start the application with $x$ threads ($x=\textit{result}$).
\end{enumerate}

\pgfplotstableread[col sep=&, header=true]{
parliament & quality
Baden-Württemberg & 93.15
\textbf{Bayern} & 89.92
Bremen & 94.05
Bundesrat & 94.53
Hessen & 94.48
Mecklenburg-Vorpommern & 95.01
Niedersachsen & 94.70
Nordrhein-Westfalen & 95.10
\textbf{Nationalrat (AT)} & 88.56
Rheinland-Pfalz & 94.34
Saarland & 95.05
Sachsen & 95.54
Thüringen & 94.21
}\ocrtable

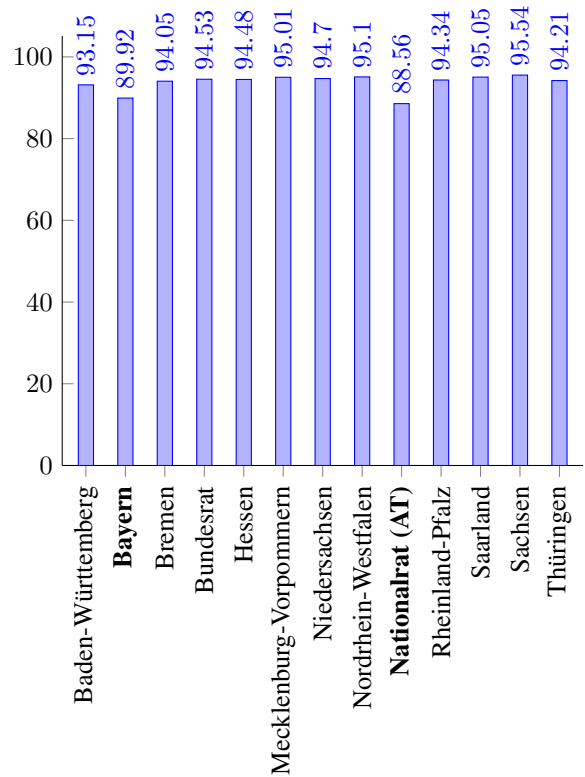
\begin{figure}[hb!]
    \begin{tikzpicture}
        \begin{axis}[
                ybar,
                axis y line*=left, 
                axis x line*=bottom,
                enlarge x limits=0.05,
                xtick=data,
                xticklabels from table={\ocrtable}{parliament},
                xticklabel style = {rotate=90, anchor=east},
                ymin=0.0,
                bar width=.2cm,
                nodes near coords,
                every node near coord/.append style={rotate=90, anchor=west}
            ]
            \addplot table[x expr=\coordindex, y=quality]{\ocrtable};
        \end{axis}
    \end{tikzpicture}
    \caption{Testing OCR quality based on \textsc{Tesseract}. Bold face refers to {Fraktur}. $y$-axis shows the percentage of correct tokens to the total number of tokens (exclude the unknown tokens).}
    \label{fig:OCRgood}
\end{figure}

\begin{table*}[ht]
\begin{tabularx}{\linewidth}{X|c|c|c|c|c|c}
\textbf{Par\-liament} &  \textbf{Period} & \rotatebox{90}{\textbf{good quality}} & \rotatebox{90}{\textbf{unknown good quality}} & \rotatebox{90}{\textbf{unknown words $\%$}} & \rotatebox{90}{\textbf{right words $\%$}} & \rotatebox{90}{\textbf{wrong words $\%$}}\\ \hline 

Baden Württem\-berg	& 1985-06-05--1996-02-08 & 	93.15\%	 & 	87.52\%	 & 	6.05\%	 & 	87.52\%	 & 	6.43\%  \\ \hline
    \textbf{Bayern} & 1946-12-16--1950-11-20	 & 	89.92\%	 & 	86.60\%	 & 	3.70\%	 & 	86.60\%	 & 	9.70\%  \\ \hline
    Bremen	 & 1967-11-08--1995-09-05  & 	94.05\%	 & 	88.73\%	 & 	5.66\%	 & 	88.73\%	 & 	5.62\% \\ \hline
    Bundesrat	& 1949-09-07--1996-12-21  & 	94.53\%	 & 	86.60\%	 & 	8.39\%	 & 	86.60\%	 & 	5.02\% \\ \hline
    Hessen	 & 1946-12-19--1998-12-16  & 	94.48\%	 & 	88.86\%	 & 	5.95\%	 & 	88.86\%	 & 	5.19\%  \\ \hline
    Mecklen\-burg-Vor\-pommern	 & 1990-10-26--2002-06-27 & 	95.01\%	 & 	88.44\%	 & 	6.92\%	 & 	88.44\%	 & 	4.64\%  \\ \hline
    Nieder\-sachsen	& 1982-06-22--1998-02-19  & 	94.70\%	 & 	88.56\%	 & 	6.47\%	 & 	88.56\%	 & 	4.96\% \\ \hline
    Nordrhein Westfalen	 & 1947-05-19--2005-04-21 & 	95.10\%	 & 	89.18\%	 & 	6.23\%	 & 	89.18\%	 & 	4.59\%  \\ \hline
    \textbf{National\-rat (AT)}	 & 1918-10-21--1930-07-16 & 	88.56\%	 & 	85.15\%	 & 	3.84\%	 & 	85.15\%	 & 	11.01\%  \\ \hline
    Rheinland\-Pfalz	& 1947-06-04--2006-02-17  & 	94.34\%	 & 	88.30\%	 & 	6.41\%	 & 	88.30\%	 & 	5.30\%  \\ \hline
    Saarland	 & 1994-09-11--1999-08-25 & 	95.05\%	 & 	89.44\%	 & 	5.91\%	 & 	89.44\%	 & 	4.65\%  \\ \hline
    Sachsen	& 1990-10-27--2004-06-25  & 95.54\%	 & 	89.17\%	 & 	6.67\%	 & 	89.17\%	 & 	4.16\% \\ \hline
    Thüringen & 1990-10-25--1994-08-09	& 	94.21\%	 & 	87.61\%	 & 	7.01\%	 & 	87.61\%	 & 	5.38\%  \\ \hline

\end{tabularx}
    \caption{
    Testing OCR quality based on \textsc{Tesseract}. Bold face refers to {Fraktur}.
    }
    \label{tab:table_spellchecking}
\end{table*}

PDFs that contain Fraktur are a challenge for OCR.
For this reason, we rescaled, eroded, and dilated them and tried to reduce noise with a filter to improve extraction, as recommended by \textsc{Tesseract}.\footnote{\url{https://tesseract-ocr.github.io/tessdoc/ImproveQuality.html}}
Figure~\ref{fig:OCRgood} shows the results of testing the OCR output quality. A spell checker was used for this test. 
Bold face columns concern extractions in Fraktur.
Most of the quality outputs are close to equal at around 94\%.
For spell checking, we used \textit{SymSpell}.
For this we used the Python library \textit{sysmspellpy}~\cite{symspellpy:2018}.
We checked every token which consists of letters or is a combination of numbers and letters.
Otherwise it was skipped, because \textit{SymSpell} processes only words or word-like tokens.
\textit{SymSpell} has three possible outputs in our case:
\begin{enumerate}
    \item The input and the output are equal to each other (which increases the number of correct words).
    \item The input and the output are \textbf{unequal} to each other (which increases number of wrong words).
    \item The output is empty; in this case \textit{SymSpell} cannot correct the input (which increases the number of unknown words).
\end{enumerate}
Moreover, it should be noted that \textit{good quality} says nothing about the number of unknown words and that \textit{unknown good quality} contains all words that are not skipped.
However, Table~\ref{tab:table_spellchecking} illustrates that the number of unknown words is significantly lower than the number of correct words.
The percentages of the numbers of correct, wrong and unknown words are based on all words, which are not skipped.
For this reason, \textit{unknown good quality} is equal in percentage to the percentage of correct words.
The National Council has the worst quality score (88.30$\%$ -- unknown good quality) and Sachsen/Saxony the best one (95.54$\%$ -- good quality).
Our test shows that OCR is sufficiently good to support NLP based on \GerParCor.

With the preprocessed version of \GerParCor it is possible to create different subcorpora to support different research endeavors:
\begin{itemize}
    \item one can use \GerParCor as a whole,
    \item without OCR-based documents,
    \item only with OCR-based documents,
    \item or only with those documents based on {Fraktur}.
\end{itemize}

In particular, we expect time-related approaches (concerning studies of language change); 
but also analyses of political language should become possible with these data on a scale that encompasses parliamentary texts from several parliaments and, at the same time, several countries.

\section{Future Work} 
Once the basic corpus has been created, it must be ensured that new releases of parliamentary minutes are continually added to the corpus.
This requires automated retrieval of the protocols and their processing.
In addition, a web-based search portal is needed to search and extract the minutes in different subsets and different formats. 
To enable this, the \textit{UIMADatabaseInterface}~\cite{Abrami:Mehler:2018} can be used, which enables storage and retrieval of UIMA documents with a number of data and document-based database systems.
Moreover, for improving the quality of OCR recognition, it is planned to train a model capable of reconstructing unknown words, which should be possible given words and their contexts.
Finally, \GerParCor should be extended to include other parliamentary documents as listed in section~\ref{sec:introduction}.

\section{Conclusion} 
We presented, \GerParCor, the currently largest German-language corpus for parliamentary protocols. 
It includes the protocols of parliaments in Austria, Germany, Liechtenstein and Switzerland.
For this purpose, the online available minutes of federal parliaments (for Germany also for state parliaments) were automatically extracted, OCRed and preprocessed with spaCy3.
Since some protocols were only available as scans, some in Fraktur, they were converted with the help of \textsc{Tesseract}.
The complete corpus with its annotations and all the programs created for it are available via GitHub.

\begin{figure*}[hb]
\begin{lstlisting}[language=XML,linewidth=\textwidth]
<annotation2:DocumentAnnotation xmi:id="23" sofa="1" dateDay="11" subtitle="17.Wahlperiode__1.Sitzung" dateMonth="5" dateYear="2021" timestamp="1620691200000"/>
<type4:DocumentMetaData xmi:id="33" sofa="1" begin="0" end="48634" language="de" documentTitle="Landtag von Baden-Württemberg-Plenarprotokoll vom 11.05.2021" documentId="Plenarprotokoll_17_1_11.05.2021_S._1-13.xmi.gz" documentUri="file:/resources/corpora/parlamentary_germany/BadenWuertemberg/xmi/17/Plenarprotokoll_17_1_11.05.2021_S._1-13.xmi.gz" documentBaseUri="file:/resources/corpora/parlamentary_germany/" isLastSegment="false"/>

<type6:Sentence xmi:id="757" sofa="1" begin="2733" end="2841"/>

<type6:Lemma xmi:id="284068" sofa="1" begin="2733" end="2748" value="Alterspräsident"/>
<type6:Lemma xmi:id="284080" sofa="1" begin="2749" end="2757" value="Winfried"/>
<type6:Lemma xmi:id="284092" sofa="1" begin="2758" end="2769" value="Kretschmann"/>
<type6:Lemma xmi:id="284104" sofa="1" begin="2769" end="2770" value=":"/>
<type6:Lemma xmi:id="284116" sofa="1" begin="2771" end="2776" value="Meine"/>
<type6:Lemma xmi:id="284128" sofa="1" begin="2777" end="2781" value="sehr"/>
<type6:Lemma xmi:id="284140" sofa="1" begin="2782" end="2791" value="verehren"/>
<type6:Lemma xmi:id="284152" sofa="1" begin="2792" end="2797" value="Dame"/>
<type6:Lemma xmi:id="284164" sofa="1" begin="2798" end="2801" value="und"/>
<type6:Lemma xmi:id="284176" sofa="1" begin="2802" end="2808" value="Herr"/>

<type6:Token xmi:id="19853" sofa="1" begin="2733" end="2748" lemma="284068" pos="178914" order="0"/>
<type6:Token xmi:id="19873" sofa="1" begin="2749" end="2757" lemma="284080" pos="178927" morph="389238" order="0"/>
<type6:Token xmi:id="19893" sofa="1" begin="2758" end="2769" lemma="284092" pos="178940" morph="389268" order="0"/>

<morph:MorphologicalFeatures xmi:id="389238" sofa="1" begin="2749" end="2757" gender="Masc" number="Sing" case="Nom" value="Case=Nom|Gender=Masc|Number=Sing"/>

<dependency:Dependency xmi:id="606959" sofa="1" begin="2733" end="2748" Governor="19893" Dependent="19853" DependencyType="PNC" flavor="basic"/>
<dependency:Dependency xmi:id="606974" sofa="1" begin="2749" end="2757" Governor="19893" Dependent="19873" DependencyType="PNC" flavor="basic"/>
\end{lstlisting}

    \caption{
    Excerpt from an annotated XMI document:
    Line 1 and 2 shows meta information from the minutes of the Baden-Würtemberg state parliament on 05/11/2021.
    This contains the title (2) as well as the date and a subtitle (1).
    For this protocol, the sentence \enquote{Alterspräsident Winfried Kretschmann: Meine sehr verehrten Damen und Herren, liebe Kolleginnen und Kollegen!} (English: \enquote{Senior President Winfried Kretschmann: Ladies and gentlemen, dear colleagues!}) is shown here in XMI.
    Line 4 shows the sentence annotation and lines 6 - 15 an exerpt of the lemma annotations; and lines 17 - 19 an excerpt from the token annotations.
    Within the serialization of the CAS document (XMI) references can be recognized, which are specified via the ID's of the respective attributes. 
    In line 21 the morphological annotation is given for the token in line 17.
    Lines 23 and 24 show an excerpt of the dependency annotations for the sentence.
    }
    \label{fig:xmi}
\end{figure*}

\section{Bibliographical References}\label{reference}

\bibliographystyle{lrec2022-bib}
\bibliography{arxiv}

\end{document}